\documentclass[10pt,twocolumn,letterpaper]{article}

\usepackage[pagenumbers]{iccv} 

\usepackage{microtype}
\usepackage{graphicx}
\usepackage{multirow}
\usepackage{booktabs}
\usepackage{colortbl}
\usepackage{subfloat}
\usepackage{floatrow}
\usepackage{float}
\usepackage{pifont}

\newfloatcommand{capbtabbox}{table}[][\FBwidth]
\newfloatcommand{capbfigbox}{figure}[][\FBwidth]

\usepackage{algorithm}
\usepackage{algorithmicx}
\usepackage[noend]{algpseudocode}

\usepackage{pifont}
\usepackage{bbding}
\usepackage{caption}
\definecolor{iccvblue}{rgb}{0.21,0.49,0.74}
\usepackage[pagebackref,breaklinks,colorlinks,allcolors=iccvblue]{hyperref}

\def\paperID{14305} 
\def\confName{ICCV}
\def\confYear{2025}

\title{GTR: \underline{G}uided \underline{T}hought \underline{R}einforcement Prevents Thought Collapse\\in RL-based VLM Agent Training}

\author{
Tong Wei$^{1,2}$\thanks{Equal contribution.}\;\thanks{Work done during an internship at Tencent. Code can be found \href{https://github.com/weit123/GTR}{here}.}
\quad Yijun Yang$^{2*}$
\quad Junliang Xing$^{1\ddag}$
\quad Yuanchun Shi$^{1}$
\quad Zongqing Lu$^{3}$
\quad Deheng Ye$^{2}$\thanks{Corresponding authors.} \\[0.5mm]
$^{1}$Tsinghua University \quad $^{2}$Tencent \quad $^{3}$Peking University\\
\tt\small wt22@mails.tsinghua.edu.cn, yijun.steven.yang@gmail.com, jlxing@tsinghua.edu.cn, \\ \tt\small shiyc@tsinghua.edu.cn, zongqing.lu@pku.edu.cn, dericye@tencent.com
}

\begin{document}

\maketitle
\begin{abstract}
Reinforcement learning with verifiable outcome rewards (RLVR) has effectively scaled up chain-of-thought (CoT) reasoning in large language models (LLMs). Yet, its efficacy in training vision-language model (VLM) agents for goal-directed action reasoning in visual environments is less established. This work investigates this problem through extensive experiments on complex card games, such as 24 points, and embodied tasks from ALFWorld. We find that when rewards are based solely on action outcomes, RL fails to incentivize CoT reasoning in VLMs, instead leading to a phenomenon we termed \textbf{thought collapse}, characterized by a rapid loss of diversity in the agent's thoughts, state-irrelevant and incomplete reasoning, and subsequent invalid actions, resulting in negative rewards. To counteract thought collapse, we highlight the necessity of process guidance and propose an automated corrector that evaluates and refines the agent's reasoning at each RL step. This simple and scalable GTR (\textbf{G}uided \textbf{T}hought \textbf{R}einforcement) framework trains reasoning and action simultaneously without the need for dense, per-step human labeling. Our experiments demonstrate that GTR significantly enhances the performance and generalization of the LLaVA-7B model across various visual environments, achieving 3-5 times higher task success rates compared to SoTA models with notably smaller model sizes.
\end{abstract}
\section{Introduction}
\label{sec:intro}
The rapid evolution of large language models (LLMs) and vision-language models (VLMs) has significantly enhanced machines' ability to comprehend general text and images. Through a comprehensive reinforcement learning (RL) training pipeline, these models can be improved based on evaluations (i.e., rewards) of their outcomes, thereby aligning with human values, emerging reasoning capabilities via long-chain-of-thought (CoT), and achieving higher success rates across various tasks. Yet, few studies have explored how to train VLM agents in dynamic visual environments to infer a sequence of correct actions based on their perceived information and ultimately accomplish specific goals. 

Although prior work \cite{zhai2025fine} has demonstrated the feasibility of RL with verifiable outcome rewards (RLVR) for fine-tuning VLM agents to solve multi-step decision-making tasks, their progress remains limited in environments involving longer episodes, larger state spaces, and more substantial reasoning requirements. In experiments conducted on representative tasks, such as the 24 points card game \cite{zhai2025fine} and the embodied environment ALFWorld \cite{shridhar2020alfworld}, we identified the bottleneck that restricts the further emergence of the model's decision-making and reasoning capabilities, which we refer to as \textbf{thought collapse}. In RL training, rewards are entirely derived from the final action, without considering the intermediate reasoning thoughts (see Fig.~\ref{fig:thought_collapse} for more details). When facing challenging and complex tasks, the guiding effect of action rewards is primarily weakened. The agent's CoT process rapidly loses diversity, resulting in incorrect, incoherent, and rigid thoughts, even with different visual and textual inputs, ultimately leading to erroneous actions and negative rewards. Although the agent continues to generate reasoning thoughts, it has already lost the ability to think, hindering the emergence of its full potential, as illustrated by the checkpoints \ding{204} and \ding{205} in Fig.~\ref{fig:thought_collapse}.

In this paper, we highlight that process guidance is critical in mitigating thought collapse during the RLVR training of VLM agents. We propose \textbf{G}uided \textbf{T}hought \textbf{R}einforcement (\textbf{GTR}), a simple and scalable framework that boosts the decision-making capabilities of agents during RL training by combining automatic thought correction and the RL-based optimization of both the agent's thoughts and actions. Compared to conventional approaches such as training a process reward model \cite{lightman2023let, uesato2022solving} or employing external verifier rewards \cite{zhang2024generative, gao2024llm, xia2024evaluating, hou2025advancing, yeo2025demystifying}, \textbf{our framework does not rely on meticulous human expert annotations or additional training, but provides more informative process supervision while preserving the flexibility of RLVR}, therefore resulting in a more efficient and effective solution for training versatile VLM agents in a variety of visual environments. \looseness-1

Specifically, as shown in Fig.~\ref{fig:gtr}, we design a plug-and-play VLM corrector model built upon any off-the-shelf VLM to evaluate and refine the agent's thoughts at each RL training step, thus automating the correction of collapsed thought trajectories. Inspired by previous works that integrate guidance into RL training \cite{levine2013guided, fujimoto2021minimalist, setlur2024rewarding}, our GTR framework allows the agent to perform SFT thought-cloning alongside PPO updates, ensuring both the rationality of the reasoning process and the correctness of the final actions. Furthermore, we address the issue of output format degradation with format rewards and repetition penalties, enhance the accuracy and coherence of thought correction through appropriate tool usage, and mitigate distribution shift in the thought-cloning process by a prevalent imitation learning method, Dataset Aggregation (DAgger) \cite{ross2011reduction}. The integration of thinking process guidance enables the VLM agent to generate a structured and reliable reasoning and action trajectory, resulting in more transparent and interpretable decision-making in complex and challenging visual environments.

Empirically, we apply GTR to train an LLaVA-7B model \cite{liu2023llava, liu2023improvedllava, liu2024llavanext}. On the highly challenging card games such as 24 points, GTR achieved over \textbf{300\% task success rate} and significantly higher returns compared to the state-of-the-art (SoTA) methods (including the agents driven by Qwen2-VL, Gemini, and GPT-4o), demonstrating that combining thought guidance with RL unleashes the decision-making potential of VLM agents more effectively. Additionally, in experiments on the embodied environment ALFWorld, GTR exhibits a better success rate and sample efficiency, proving the generality of our framework across diverse tasks.

\section{Related Works}
\label{sec:related_works}

\subsection{LLMs/VLMs as Decision-making Agents}
Recent advancement in foundation models has expanded their applications beyond text and image generation to complex decision-making tasks, focusing on their ability to reason, plan, and interact with dynamic environments \cite{xi2025rise, yang2023foundation,li2024muep,yang2024embodied}. Many previous works have leveraged prompting techniques to activate decision-making capabilities from frozen LLMs \cite{wei2022chain, yao2023tree, yao2023react, wang2023describe, wang2023voyager, huang2022language, park2023generative, ahn2022can, shinn2023reflexion, wu2023autogen}. Furthermore, agents with VLMs as backbones can perform reasoning based on visual inputs, either by translating their observations to text descriptions or aligning their embeddings with LLMs \cite{ahn2022can, driess2023palm, gao2024physically, huang2023voxposer, mu2023embodiedgpt, sumers2023distilling, yang2024octopus, brohan2023rt,zhou2024wall}. However, these approaches either do not involve model adjustments or are limited to pretraining additional layers on fixed datasets, making them unable to adapt to dynamic environments through interaction. In contrast, we explore reinforcement learning to fine-tune the entire VLM, enhancing the model's ability to reason and make decisions by interacting with the environment through open-ended text.

\subsection{RL Training for LLMs/VLMs}
Reinforcement learning has been proven to be an effective means of incentivizing the emergence of capabilities in LLMs and VLMs. Some studies focus on RL from human feedback (RLHF), which leverages human feedback datasets to learn a reward model and then conducts reinforcement learning \cite{ziegler2019fine, stiennon2020learning, ouyang2022training, achiam2023gpt, openaigpt4o, team2024gemini, sun2023aligning}. Another common approach involves using human preference data to align the model with humans by modifying the reward function of the RL algorithm \cite{ramamurthy2022reinforcement, carta2023grounding, zhou2024archer, rafailov2023direct}. The distinction of our study lies in conducting RL finetuning directly based on rewards provided by the environment.

Recently, methods utilizing long Chain-of-Thought and inference-time scaling have demonstrated superior slow-thinking capabilities \cite{openaio1, guo2025deepseek}. Techniques such as process reward models \cite{lightman2023let, uesato2022solving, wang2023math}, searching algorithms \cite{feng2023alphazero, trinh2024solving, xin2024deepseek}, and GRPO \cite{shao2024deepseekmath} have achieved breakthroughs in tasks such as solving mathematical problems. In contrast, our work focuses on fast-thinking decision-making for the entire action sequences, aiming to achieve specific goals in interactive environments. Additionally, our approach incorporates multimodal information, integrating vision-language reasoning that expands its applicability to a broader range of tasks.

Our research is closely related to RL4VLM \cite{zhai2025fine}, which proposes a method for directly finetuning VLMs using RL. However, as we discuss later, while this approach significantly improves more straightforward tasks, its performance gains are limited in more complex environments.

\subsection{Process Guidance Approaches}
Extensive studies on deep-thinking LLMs in mathematical reasoning have affirmed the contribution of process supervision to enhancing the logical coherence of model outputs. Approaches can be broadly categorized into three types. The first involves training a Process Reward Model (PRM) to assess the reasoning process \cite{lightman2023let, uesato2022solving}, but this method requires costly human annotations to obtain high-quality data. The second approach uses Monte Carlo estimation \cite{wang2023math, luo2024improve, chen2024alphamath} or credit assignment techniques \cite{wang2024q, cui2025process, yuan2024free} to infer the quality of thoughts from outcomes. While effective for deep-thinking tasks with long intermediate steps, this method is less suitable for the sequential decision-making problems of agents with single-step reasoning. Lastly, methods such as LLM-as-a-judge \cite{zhang2024generative, gao2024llm, xia2024evaluating} or length-based rewards \cite{hou2025advancing, yeo2025demystifying} can directly evaluate reasoning quality. However, in complex tasks, the numerical rewards provided by these methods often lack sufficient information to guide RL training effectively.
\section{Thought Collapse}
\label{sec:thought_collapse}


Unlike purely text-based LLM agents, due to the integration of multimodal information and the increased complexity of the decision-making process, training VLM agents in interactive visual environments using RL raises more challenging issues. Previous research has attempted to establish an algorithmic framework that demonstrates the potential of RL training to unleash the decision-making capabilities of VLMs in simple tasks, also highlighting the crucial role of Chain-of-Thought (CoT) reasoning \cite{zhai2025fine}. However, in some complex environments such as the 24-point game and the embodied household tasks from ALFWorld, RL training has shown limited improvement, with no significant emergence performance observed.

Our experiments may identify the root of this problem: the agent's thought process often becomes irrational or templated during RL training, significantly impairing its decision-making abilities. As illustrated in Fig.~\ref{fig:thought_collapse}, the agent fails to improve task success rates or episode returns. Instead, it generates fragmented thoughts and degenerates into a strategy of producing similar outputs for different states. Although the agent continues to output thoughts and action decisions at this stage, it has clearly lost its ability to engage in meaningful reasoning and decision-making. We term this catastrophic phenomenon as ``thought collapse''.

\begin{figure}[htbp]
    \vspace{-0.5em}
    \centering
    \includegraphics[width=0.8\linewidth]{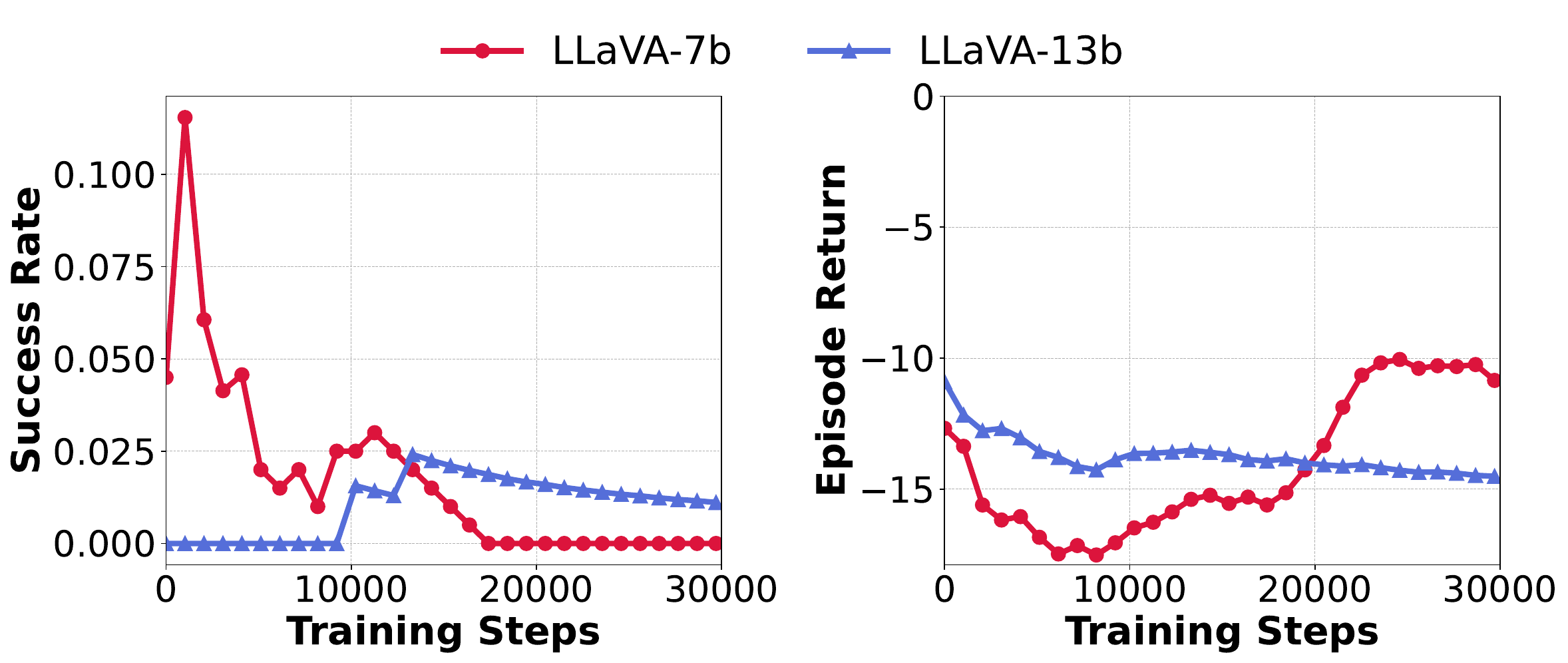}
    \caption{\textbf{Thought collapse persists for larger model scales and training budgets.} We train LLaVA 7B and 13B models for 30k RL steps but still observe the degraded performance.}
    \vspace{-0.5em}
    \label{fig:model_size_training_duration}  
\end{figure}

We first investigated whether these observations stemmed from insufficient base model capabilities or training budgets. To this end, we conducted RL training on two model scales: LLaVA-v1.6-mistral-7B and LLaVA-v1.6-vicuna-13B, and extended the training steps from 15k to 30k. The results in Fig.~\ref{fig:model_size_training_duration} indicate that models of different scales exhibit similar trends of thought collapse, and extending training duration does not mitigate the issue.

Thus, we assume that thought collapse arises from RL training itself, in which rewards are determined entirely by the final actions generated by the agent. As a result, the thought process - longer and more fundamental than action output - remains unevaluated and unsupervised. This issue becomes particularly pronounced in tasks with longer episodes, larger state spaces, and greater complexity, ultimately causing the training process to deviate from its intended reasoning trajectory due to accumulated errors. Therefore, we suggest that process guidance can be pivotal in preventing thought collapse during the RL training of VLM agents, and propose a novel method, Guided Thought Reinforcement (GTR), built upon the existing RL framework to counteract this problem.

\section{Guided Thought Reinforcement}
\label{sec:gtr}

\begin{figure*}[t]
  \centering
  \includegraphics[width=0.99\linewidth]{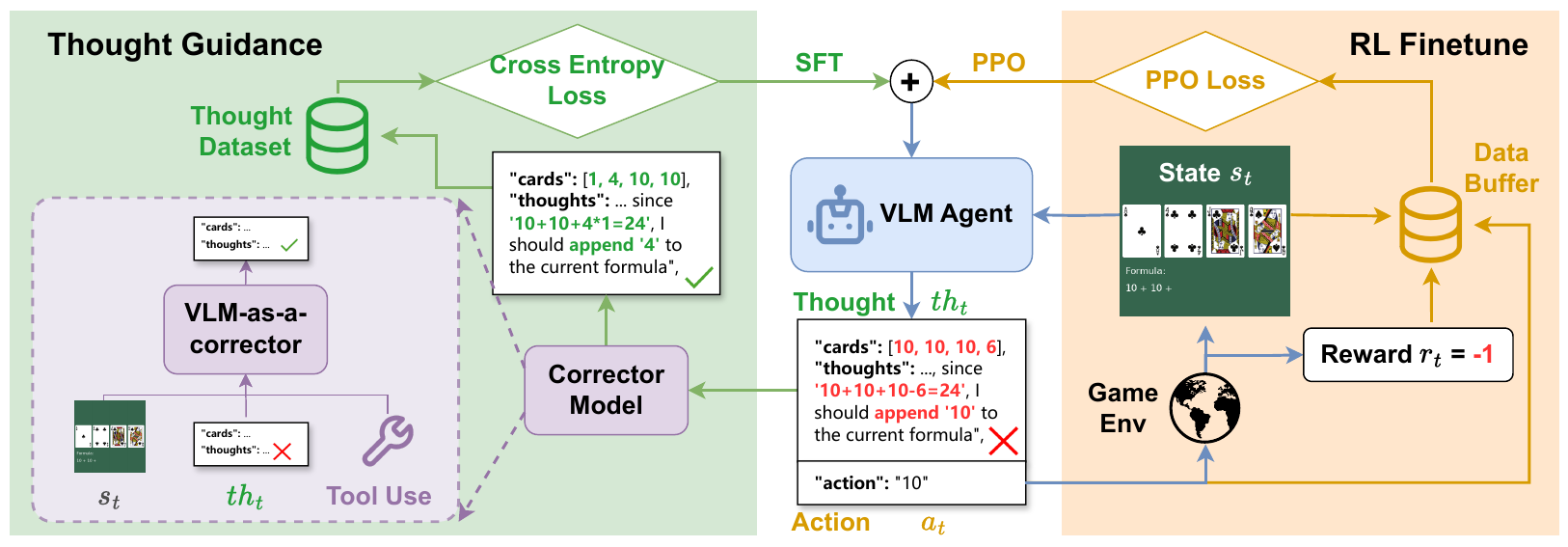}
  \caption{\textbf{Overview of the GTR framework.} We modify the RL finetuning (orange region) of VLM agents by introducing automated thought correction (green region) as guidance, leveraging an off-the-shelf VLM model as a corrector (purple region). GTR performs SFT updating for the agent's thought tokens and PPO updating for its action tokens, thereby training thoughts and actions simultaneously.}
  \label{fig:gtr}
\end{figure*}

\subsection{RL Training of VLM Agents}
Our backbone RL algorithm adopts the well-established framework that uses PPO to finetune VLMs for sequential decision-making \cite{schulman2017proximal, zhai2025fine}. The post-processing of the policy function extracts the keyword ``$\mathtt{action:}\ a$'' from the VLM's text outputs as the action. If the output text does not contain this keyword, the agent performs random exploration, selecting randomly from a set of all legal actions. Formally, given the VLM's output $\boldsymbol{v}^{\text{out}}$ and the set of legal actions $\mathcal{A}$, the post-processing function $f$ is defined as:
\begin{equation}
\label{eqn:postprocessing}
    f(\boldsymbol{v}^{\text{out}}) =
    \begin{cases}
        \textit{a}, & \text{if ``} \mathtt{action: \textit{a}}\text{''} \in \boldsymbol{v}^{\text{out}} \\
        \mathtt{Unif}(\mathcal{A}), & \text{otherwise}
    \end{cases}
\end{equation}

The action probability required for policy gradient is calculated from the generation probabilities of each token in the output text. Additionally, a scaling factor is employed to balance the longer length of CoT outputs compared to action outputs. If $\pi_\theta$ denotes the policy, $o_t$ and $a_t$ represent the observation and action, $\boldsymbol{v}_t^{\text{in}}$ as input tokens, $\boldsymbol{v}_t^{\text{tht}}$ and $\boldsymbol{v}_t^{\text{act}}$ represent CoT reasoning tokens and action tokens respectively, $[:i]$ denotes the first $i$ tokens, this calculation is shown as:
\begin{align}
    &\log \pi_\theta(a_t|o_t, \boldsymbol{v}_t^{\text{in}}) \label{eqn:loglikelihood}\\
    = &\lambda \log \pi_\theta(\boldsymbol{v}_t^{\text{tht}}|o_t, \boldsymbol{v}_t^{\text{in}}) + \log \pi_\theta(\boldsymbol{v}_t^{\text{act}}|o_t, \boldsymbol{v}_t^{\text{in}}, \boldsymbol{v}_t^{\text{tht}}) \nonumber \\
    = &\lambda \sum\log\frac{p(o_t, \boldsymbol{v}_t^{\text{in}}, \boldsymbol{v}^{\text{tht}}_{[:i]})}{p(o_t, \boldsymbol{v}_t^{\text{in}}, \boldsymbol{v}^{\text{tht}}_{[:i-1]})} + \sum\log\frac{p(o_t, \boldsymbol{v}_t^{\text{in}}, \boldsymbol{v}_t^{\text{tht}}, \boldsymbol{v}^{\text{act}}_{[:i]})}{p(o_t, \boldsymbol{v}_t^{\text{in}}, \boldsymbol{v}_t^{\text{tht}}, \boldsymbol{v}^{\text{act}}_{[:i-1]})}. \nonumber
\end{align}

Based on the action log probability and the reward feedback from the environment, the agent performs PPO updates to adjust token generation probabilities and discover the optimal policy while interacting with the environment.

\subsection{Process Guidance from a VLM Corrector}
\begin{figure}[H]
  \centering
  \vspace{-0.5em}
    \includegraphics[width=0.8\linewidth]{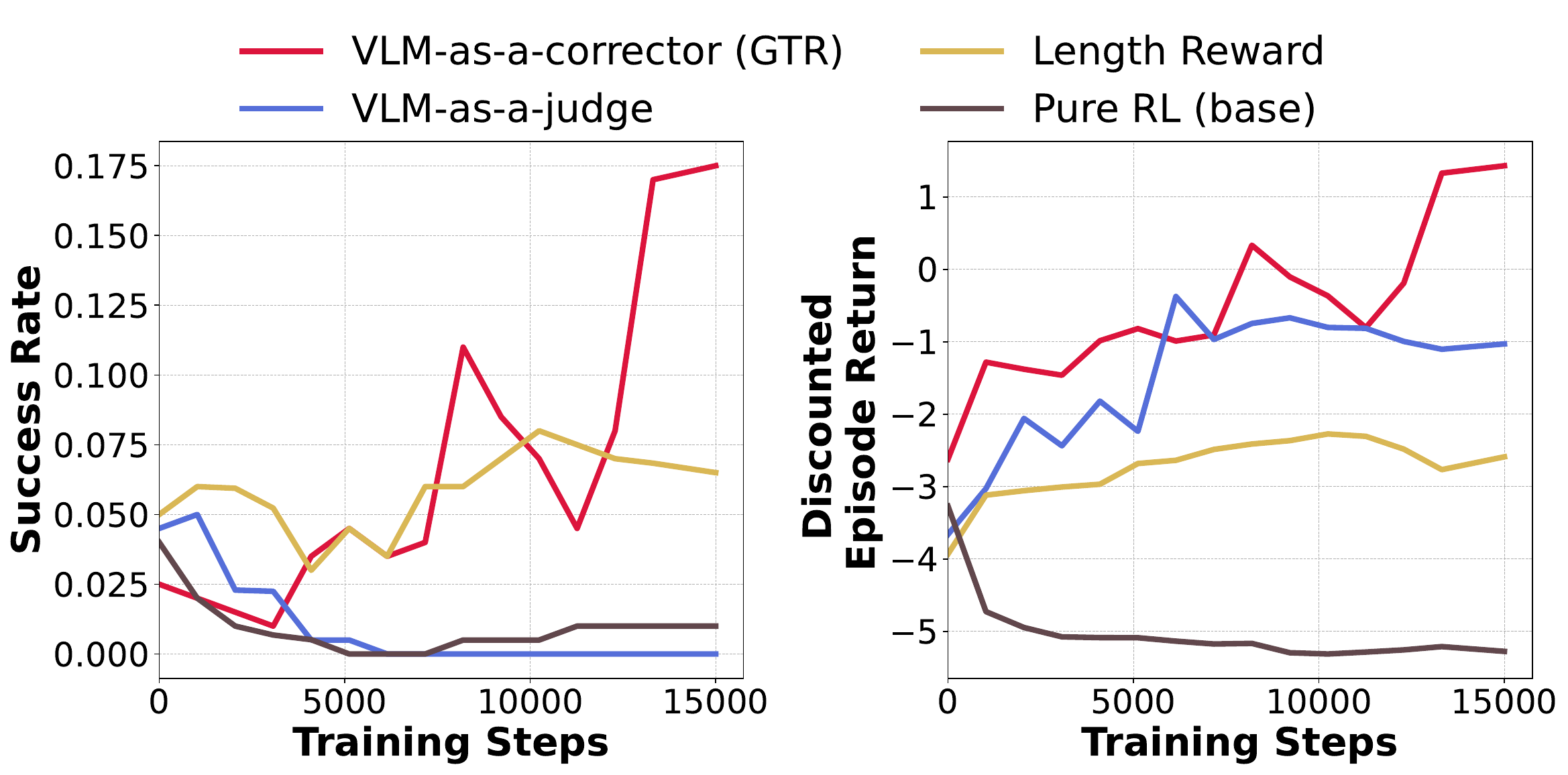}
  \caption{\textbf{Comparison among different process guidance methods in the 24 points card game.} Trivial numerical rewards provided by the VLM judge or rule-based evaluation cannot incentivize the agent's reasoning thoughts and higher success rate.}
  \vspace{-0.5em}
  \label{fig:process_guidance_approaches}    
\end{figure}

Various approaches that introduce process guidance to model training have been explored. Process Reward Models (PRMs) represent a traditional and widely used method for process supervision, particularly in mathematical reasoning tasks with LLMs \cite{uesato2022solving, lightman2023let}. However, high-quality vision-language data annotation is extremely expensive and time-consuming. Moreover, learning with static datasets cannot cover a vast variety of decision-making scenarios, often leading to biased policies when deployed to dynamic environments.
An intuitive solution to the above challenges is the VLM-as-a-judge approach, which induces a VLM to score the agent's outputs \cite{zhang2024generative, gao2024llm, xia2024evaluating}. However, this method does not demonstrate reasonable performance (see Fig.~\ref{fig:process_guidance_approaches}). While episodic returns increase to some extent, task success rates fail to improve. This is because using naive numerical scores does not provide sufficient and accurate guidance for effective RL training, especially given the strong reward-hacking capabilities of large models. In challenging tasks where the model's baseline ability is weak, the lack of positive incentives can cause the agent to fall into passive exploration, further hindering performance. Similar issues also arise with methods like length-based rewards \cite{yeo2025demystifying}, which aim to encourage longer thoughts.

\vspace{-0.5em}

These limitations call for a simple, scalable, and informative process guidance framework, and we demonstrate the corrector model as a promising solution. Our GTR framework leverages an external VLM as a corrector model, which first evaluates the agent's thought at each step for visual recognition accuracy and reasoning correctness, then refines the original thought. When the corrector identifies inconsistencies or errors, the model performs corrections based on the agent's outputs. To incorporate corrected thoughts into the agent while preserving scalability, we draw inspiration from the recent work~\cite{hu2023thought,yang2024embodied} and add a simple SFT loss over the thought tokens $\boldsymbol{v}_t^{\text{tht}}$ to the PPO loss, aligning the agent's reasoning with the corrected thought trajectories, as shown in Fig.~\ref{fig:gtr}.
Formally, if we denote the agent's thought output as $th$ and action as $a$, given agent model $\pi_\theta$ and corrector model $\pi_{\text{corr}}$, the objective of GTR can be represented as:
\begin{equation}
\label{eqn:objective_original}
    \min_{\theta} \mathop{\mathbb{E}}_{{o,(th,a)} \sim \pi_\theta}\left[\mathcal{L}_{\text{PPO}}(o, a) + \mathcal{L}_{\text{SFT}}(o, \pi_{\text{corr}}(o, th))\right],
\end{equation}
where
\begin{align}
\label{eqn:loss}
\mathcal{L}_{\text{PPO}}(o, a) &= - \min \left(\frac{\pi_\theta(a|o)}{\pi_{\theta_k}(a|o)} A^{\pi_{\theta_k}}(o,a), \right. \nonumber \\ 
& \left.\text{clip}\left({\frac{\pi_\theta(o|s)}{\pi_{\theta_k}(o|s)}, 1+c, 1-c}\right) A^{\pi_{\theta_k}}(o,a)\right), \nonumber \\
\mathcal{L}_{\text{SFT}}(o, th) &= -\sum_t \log P(th_t|o, th_{<t}; \theta).
\end{align}

The $\pi_{\text{corr}}$ eliminates the need for careful human annotations and provides more informative and direct guidance than numerical scores. Unlike behavior cloning, the correction process does not require an expert-level external model to obtain high-quality reference trajectories, which is verified by our empirical experiments in Table~\ref{tab:evaluation_points24}: our model substantially surpasses the performance of the corrector model (GPT-4o + Tool). \looseness-1

\subsection{Mitigate Distribution Shift in Thought Cloning}
Research such as TD3+BC \cite{fujimoto2021minimalist}, which also integrates supervised signals with reinforcement learning, typically focuses on the off-policy or offline RL algorithms. However, we encountered a distribution shift issue when incorporating thought cloning into the PPO training of VLM agents. As the agent's policy iteratively updates, the PPO algorithm discards previous data and resamples in each round. Performing thought cloning on this non-i.i.d dataset can lead to catastrophic forgetting or error accumulation. To address this, we adopt an interactive imitation learning algorithm, Dataset Aggregation (DAgger) \cite{ross2011reduction}, aggregating all historical corrections and samples from them. This approach has been proven to converge to the expert policy, specifically the outputs of the corrector model. If $\mathcal{B}$ represents the PPO data buffer and $\mathcal{D}$ denotes the DAgger thought dataset, Eqn.~\ref{eqn:objective_original} can be rewritten as below.
\begin{equation}
\label{eqn:objective_dagger}\min_{\theta} \mathop{\mathbb{E}}_{(s,a)\sim\mathcal{B}}\!\!\!\mathcal{L}_{\text{PPO}}(s, a)+\mathop{\mathbb{E}}_{(s,th)\sim\mathcal{D}}\!\!\!\mathcal{L}_{\text{SFT}}(s, \pi_{\text{corr}}(s, th)).
\end{equation}
To summarize, we conclude our method in Algorithm \ref{alg:gtr}.

\begin{algorithm}[t!]
\caption{Training Procedure of the Proposed GTR}
\label{alg:gtr}
	\begin{algorithmic}[1]
	    \State \textbf{Input:~}Environment 
        $\mathtt{env}$, prompt construction function $h$, post-processing function $f$, agent model $\pi_{\theta_0}$, corrector model $\pi_{\text{corr}}$, 
            \State \textbf{Input:~} Replay buffer size $B$, update epoch $K$
            \State $\mathcal{D}\leftarrow\varnothing$ \Comment{Thought cloning dataset}
            \For{$k=0$ to $K-1$}
                \State $\mathcal{B}\leftarrow\varnothing$ \Comment{On-policy data buffer}
                \State $s_t = \mathtt{env}$.reset()
                \While{$|\mathcal{B}| < B$}
                    \State $\boldsymbol{v}_t^{\text{in}} = h(o_t)$ \Comment{Obtain prompt}
                    \State $\boldsymbol{v}_t^{\text{out}} = (\boldsymbol{v}_t^{\text{tht}}, \boldsymbol{v}_t^{\text{act}}) = \pi_{\theta_k}(o_t, \boldsymbol{v}_t^{\text{in}})$
                    \State $a_t = f(\boldsymbol{v}_t^{\text{out}})$   \Comment{Obtain action with Eqn.~\ref{eqn:postprocessing}}
                    \State Compute $\log \pi_\theta(a_t|o_t, \boldsymbol{v}_t^{\text{in}})$ with Eqn.~\ref{eqn:loglikelihood}
                    \State $\boldsymbol{\hat{v}}_t^{\text{tht}} = \pi_{\text{corr}}(o_t, \boldsymbol{v}_t^{\text{tht}})$ \Comment{Thought correction}
                    \State $r_t, o_{t+1} = \mathtt{env}$.step($a_t$)
                    \State $\mathcal{B} \leftarrow \mathcal{B} \cup (o_t, a_t, r_t, \boldsymbol{v}_t^{\text{out}}, \log \pi_\theta(a_t|o_t, \boldsymbol{v}_t^{\text{in}}))$
                    \State $\mathcal{D} \leftarrow \mathcal{D} \cup (o_t, \boldsymbol{\hat{v}}_t^{\text{tht}})$
                \EndWhile
                \State Sample mini-batch $b$ from $\mathcal{B}$, $d$ from $\mathcal{D}$
                \State Compute $\mathcal{L}_{\text{PPO}}$ with $b$
                \State Compute $\mathcal{L}_{\text{SFT}}$ with $d$ \Comment{Eqn.~\ref{eqn:loss}}
                \State $\theta_{k+1} = \arg\min_\theta(\mathcal{L}_{\text{PPO}} + \mathcal{L}_{\text{SFT}})$ \Comment{Eqn.~\ref{eqn:objective_dagger}}
            \EndFor
            \State \textbf{Output:~}$\pi_{\theta_K}$
	\end{algorithmic}
\end{algorithm}

\subsection{Improving Data Quality}
We observe that RL training, which requires the model to explore and discover better outputs, also introduces the risk of generating poor outputs. During the RL fine-tuning of VLMs, it is common for the model to produce an invalid format or exhibit excessive repetition, a phenomenon also noted in other recent studies \cite{guo2025deepseek, hou2025advancing}. Therefore, we incorporate a token-level repetition penalty during the model's response generation process and explicitly integrate format judgment into the corrector model so that answers with a valid format receive a format reward at each step. These measures significantly improve the stability of the model's output format.

While general-purpose VLM correction models possess extensive general knowledge and reasoning capabilities, they may lack task-specific expertise. To this end, we leverage the function-calling ability of large models, enabling the corrector model to access specific information or function modules during the correction process. This further enhances the accuracy and credibility of corrections. For instance, in the 24-point game, the corrector may invoke a piece of Python code that calculates possible equations that evaluate to 24.
The introduction of corrector models also enables data sampling control. By referencing the corrector's judgments, we can truncate episodes when the agent enters unsolvable or meaningless states, thereby improving the data quality in the buffer and assisting RL training efficiency. In the 24-point game, the episode ends if the agent is impossible to complete the task. In ALFWorld, we apply truncation when the agent's action sequence becomes excessively long or when it continuously repeats the same ineffective action.
\section{Experiments}
\label{sec:exps}

\subsection{Experimental Setup}
\paragraph{Environments} We mainly base our experiments on the Points24 task within the $\mathtt{gym\_cards}$ environment \cite{zhai2025fine}, which requires the model to perform fine-grained visual recognition of poker cards and then engage in language reasoning. Tasks other than Points24 require fewer recognitions, significantly smaller state and action spaces, and shorter action sequences (typically solvable within five steps). As a result, outcome rewards are sufficient to guide RL training, and thought collapse rarely occurs. In contrast, the correct action sequence of Points24 can exceed 10 steps, and the combination of a long formula and larger action space creates a much broader decision-making space. Previous methods have not achieved satisfactory results. Therefore, our experiments focus primarily on this complex problem.

At each observation $o_t$ of Points24, the agent observes an image of four poker cards and the current formula under them. For each step, the agent can choose a number in $[1,10]$ or an operator in $\{+, -, \times, /, =\}$ to append to the current formula. Only numbers appearing in the cards and operators are legal actions, with ``J'', ``Q'', and ``K'' all treated as 10. The goal is to obtain a formula that equals 24 using all the cards. The agent gets -1 as a reward for illegal actions and incorrect formulas and gets 10 as a reward when forming a correct formula. \looseness-1

To validate the generality of thought collapse and the GTR framework, we also evaluate performance on the ALFWorld benchmark, a multimodal simulation platform featuring various embodied household tasks \cite{shridhar2020alfworld}. These tasks are divided into six categories: Pick \& Place, Clean \& Place, Heat \& Place, Cool \& Place, Look in Light, and Pick Two Objects \& Place. Given the current visual observation, the agent needs to determine the next action following a predefined instruction, such as ``go to cabinet 1'', ``open drawer 1'' or ``cool apple 1 with fridge 1''. The tasks involve navigation and interaction with objects in the environment, presenting significant challenges in visual recognition, long-term planning, and commonsense reasoning. 

Notably, ALFWorld provides visual and textual observations of the environment, and previous works have utilized both as input, thereby reducing the difficulty of visual recognition. We observe with various models that this setting often leads the agent to rely heavily on text descriptions while neglecting visual input. Therefore, to evaluate the comprehensive multimodal capability of agents and to better simulate real-world scenarios, we removed the text description in our experiments. We also incorporate historical actions as part of the input, making it more suitable to assess the sequential decision-making capabilities of agents.

\paragraph{Baselines} Our primary baseline is RL4VLM \cite{zhai2025fine}, which directly uses environmental rewards for PPO training. It is the first framework to fine-tune VLM agents with RL and represents the SoTA method. Accordingly, we consider the method that solely clones the corrector's thoughts, without RL (SFT-only). When evaluating final performance, we also include the LLaVA-mistral-7B base model and commercial API-based models. For Alfworld, the challenging environment with navigation and household tasks, we collect an expert dataset using a script policy and fine-tune several SoTA VLM models based on this dataset as baselines.

\paragraph{Training Details}
The GTR framework leverages a corrector model with established capabilities for automating thought correction. In our experiment, we select the GPT-4o model for this role. Consistent with previous research, to ensure that RL training begins with a model possessing reasonable instruction following abilities, all training starts from an SFT-initialized LLaVA-mistral-7B model. We train the model for 15,000 steps on the Points24 task and 5,000 steps on the ALFWorld task, aligning with the baseline settings. On a single Nvidia (40GB) GPU, the training process with LoRA \cite{hu2022lora} takes approximately 30 hours to complete.

\subsection{Process Guidance Improves VLM Decision-making Capabilities}
\begin{figure*}[t!]
\begin{floatrow}
\ffigbox[0.60\textwidth]{%
\centering
\includegraphics[width=0.9\linewidth]{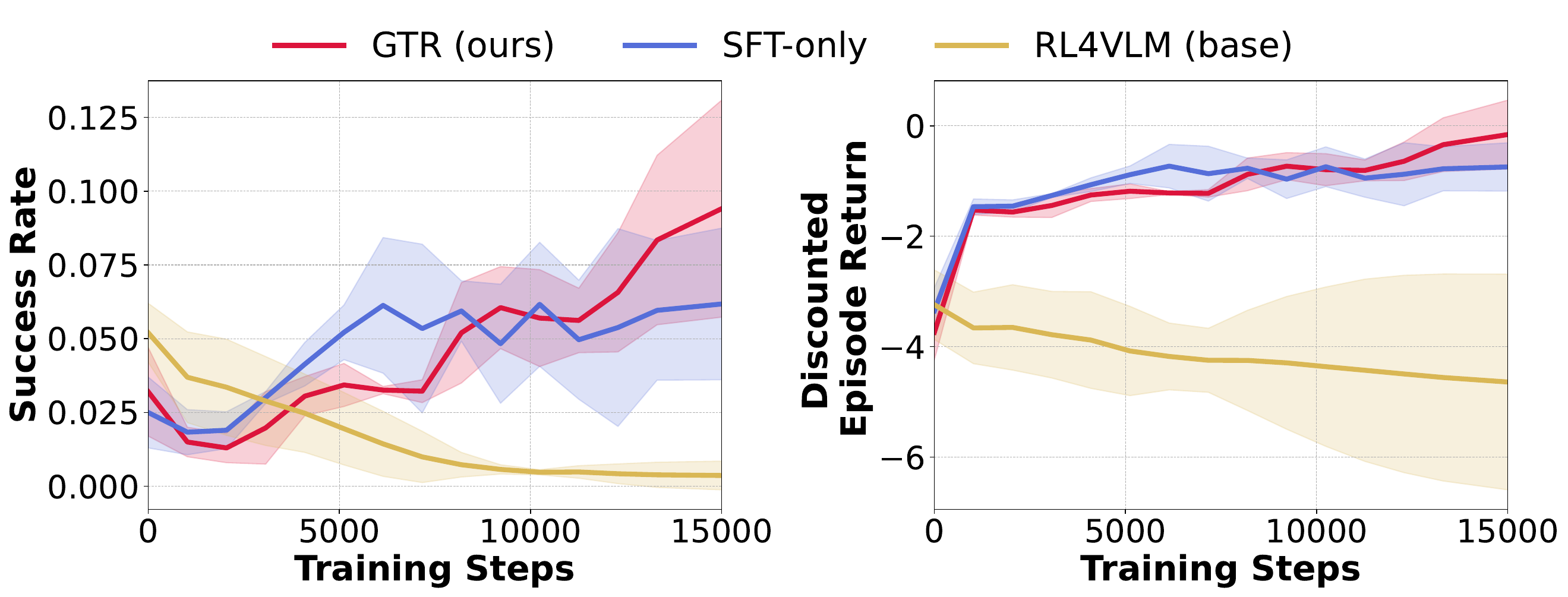}
}{%
  \caption{\textbf{Training curves on the 24 points game environment.} Compared to the baseline methods, our GTR framework, integrating process guidance with RL, achieves better performance while maintaining a rational reasoning process. Curves are smoothed for better readability. Since GTR and SFT-only employ truncation strategies, we plot $\gamma=0.9$ discounted returns in the figure for a fair comparison.}
  \label{fig:main_result_points24}
}
\hspace{-15pt}
\capbtabbox[0.38\textwidth]{%
    \centering
    \resizebox{0.7\linewidth}{!}{
        \begin{tabular}{ccc}\toprule
    \textbf{Model} & \textbf{SR(\%)}  & \textbf{ER}  \\ \midrule
    CNN+RL* & 0 & \textbf{-1.12} \\
    Gemini* & 0 & -2.68 \\
    GPT4-V* & 0 & -4.39 \\
    GPT4o & 2.5 & -6.35 \\
    GPT4o + Tool & 13.5 & -3.59 \\
    Qwen2-VL-72B* & 4.5 & / \\
    LLaVA-7b-sft & 3.0 & -15.30 \\ 
    RL4VLM & 2.5 & -12.95 \\
    SFT-only & 11.0 & -2.88 \\
    \textbf{GTR} & \textbf{17.5} & -2.17 \\ \bottomrule
    \end{tabular}
    }
}{%
  \caption{\textbf{Evaluation result of different models on the Points24 task.} GTR demonstrates significant advantages over other methods in both task success rate and episode returns. SR - success rate, ER - episode return, * - reported in previous work.}
  \label{tab:evaluation_points24}
}
\end{floatrow}
\vspace{-10pt}
\end{figure*}

As illustrated in Figure \ref{fig:main_result_points24} and Table \ref{tab:evaluation_points24}, our GTR algorithm demonstrates significant and consistent performance improvements on the Points24 task compared to existing models by a 3- to 5-fold increase in success rate and a higher return. This achievement highlights the significance of process guidance in RL training for VLMs in complex tasks. Furthermore, GTR outperforms the thought cloning method and surpasses GPT4o with tool function calling, the corrector model itself. This demonstrates that RL allows the agent to go beyond simple imitation, exploring and discovering superior strategies under the guidance of the corrector.

\begin{figure}[t]
  \centering
    \includegraphics[width=\linewidth]{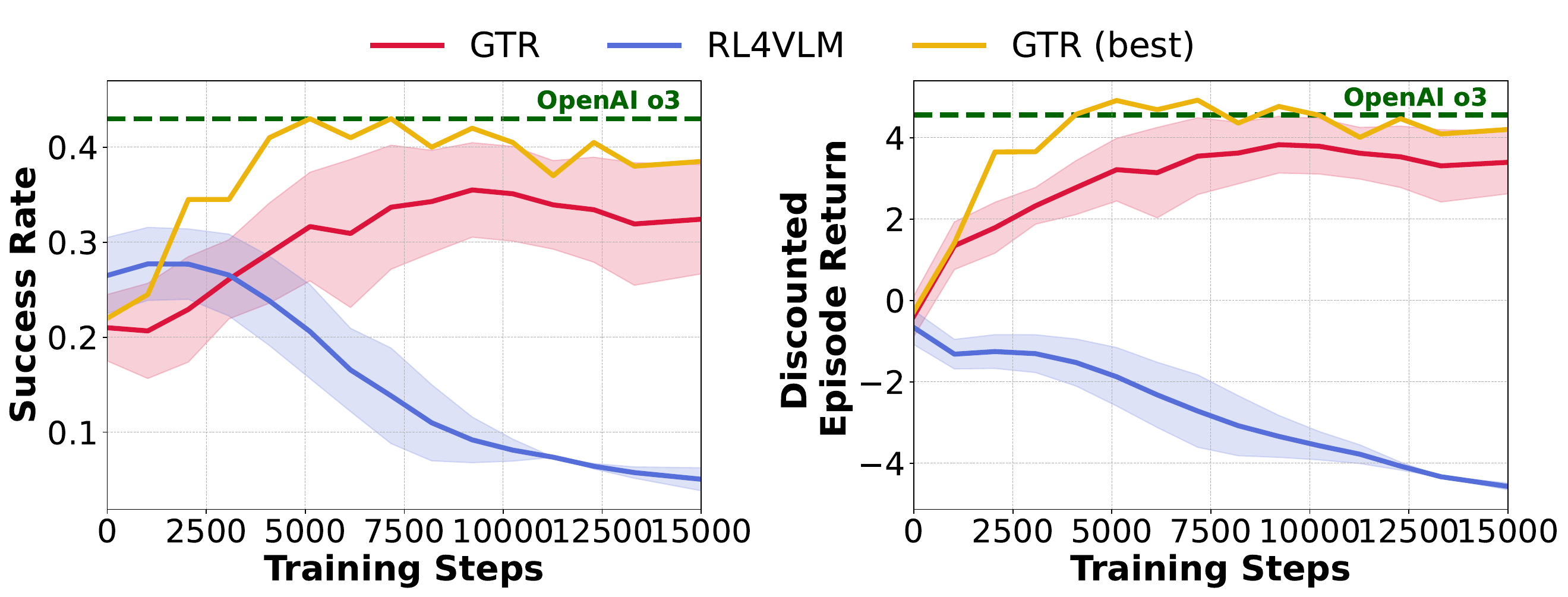}
  \caption{\textbf{Comparison of training curves on Qwen2.5-VL-7B agents.} GTR boosts model's reasoning capability, even reaching o3-level performance.}
  \label{fig:main_qwen} 
  \vspace{-0.5em}
\end{figure}

We also conducted experiments using the more recent and powerful Qwen2.5-VL-7B model. As shown in Figure \ref{fig:main_qwen}, GTR continues to deliver substantial performance improvements, even enabling the agent to reach o3-level performance within a short training period. In contrast, the RL4VLM's performance degrades as the training budget increases.


We also assess the performance of GTR on the other tasks in $\mathtt{gym\_cards}$, shown in Table \ref{tab:evaluation_other}. Due to smaller state and action spaces and significantly shorter episode lengths, thought collapse is not prominent in these tasks. Nevertheless, GTR still achieves noticeable improvements, performing comparably to Qwen2-VL, which is 10 times larger than our model in size, highlighting the effectiveness of our framework.

\begin{figure*}[t!]
\begin{floatrow}
\ffigbox[0.45\textwidth]{%
\centering
\includegraphics[width=0.9\linewidth]{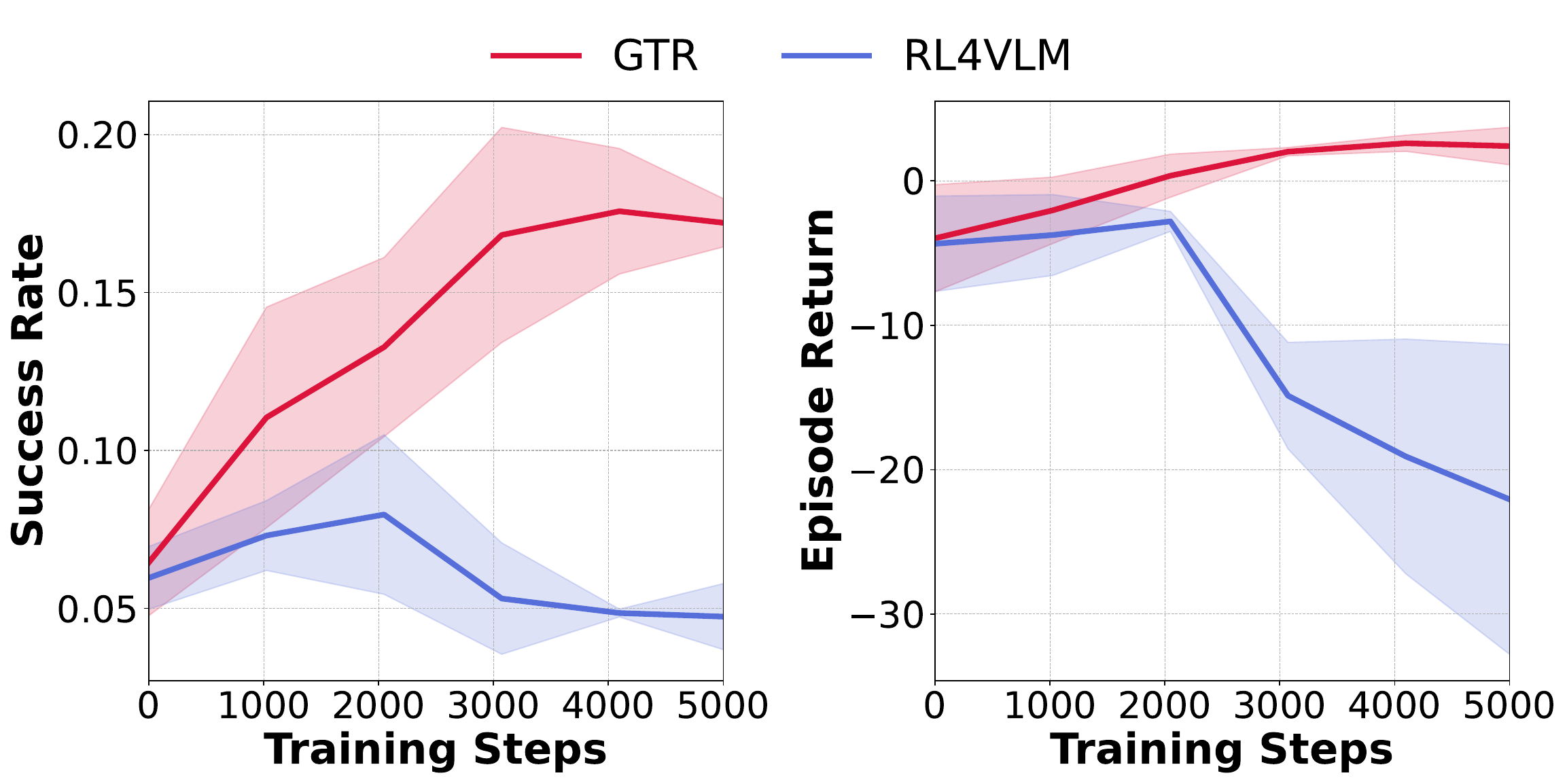}
}{%
  \caption{\textbf{Comparison of training curves between GTR and RL4VLM in the ALFWorld environment.} RL4VLM fails to facilitate model learning, leading to thought collapse effectively. GTR, however, enables the agent's performance to improve steadily, ultimately achieving superior results.}
  \label{fig:main_alfworld}
}
\hspace{-15pt}
\capbtabbox[0.55\textwidth]{%
    \centering
    \resizebox{\linewidth}{!}{
        \begin{tabular}{ccccccccc}\toprule
     \textbf{Model} & \textbf{Text Obs} & \textbf{Average} & \textbf{Pick} & \textbf{Clean} & \textbf{Heat} & \textbf{Cool} & \textbf{Look} & \textbf{Pick2} \\ \midrule
    CNN+RL* & \ding{52} & 0 & 0 & 0 & 0 & 0 & 0 & 0 \\
    Gemini* & \ding{52}& 0.14 & 0.35 & 0 & 0 & 0 & \textbf{0.16} & 0.12 \\
    GPT4-V* & \ding{52}& 0.20 & 0.38 & \textbf{0.18} & 6.7 & 0.18 & 0.12 & 0.15 \\
    LLaVA-sft* & \ding{52} & 0.18 & 0.39 & 0.14 & 0.11 & 0 & 0 & \textbf{0.29} \\ 
    RL4VLM* & \ding{52} & \textbf{0.22} & \textbf{0.47} & 0.10 & \textbf{0.14} & \textbf{0.19} & 0.15 & 0.18 \\ \hline
    MiniGPT-4 & \ding{55} & 0.16 & 0.04 & 0 & 0.19 & 0.17 & \textbf{0.67} & 0.06 \\
    BLIP-2 & \ding{55} & 0.04 & 0 & 0.06 & 0.04 & 0.11 & 0.06 & 0 \\
    LLaMA-Adapter & \ding{55} & 0.13 & 0.17 & \textbf{0.10} & \textbf{0.27} & 0.22 & 0 & 0 \\
    LLaVA-sft & \ding{55} & 0.06 & 0.14 & 0.05 & 0 & 0.06 & 0 & 0 \\
    RL4VLM & \ding{55} & 0.04 & 0.15 & 0 & 0 & 0 & 0 & 0\\
    \textbf{GTR} & \ding{55} & \textbf{0.18} & \textbf{0.37} & 0.07 & 0.08 & \textbf{0.33} & 0.23 & \textbf{0.20} \\ \bottomrule
    \end{tabular}
    }
}{%
  \caption{\textbf{Comparison of success rates across different models in the ALFWorld environment.} We present the peak performance in the training curve for RL methods. \ding{52}/\ding{55} denotes whether the environment gives textual descriptions of the current observation alongside visual images, which assists the agent's decision-making. * - reported in previous work.}
  \label{tab:evaluation_alfworld}
}
\end{floatrow}
\vspace{-10pt}
\end{figure*}

In Figure \ref{fig:main_alfworld}, we compare the training curves of GTR and RL4VLM on the ALFWorld embodied environment. The results show that RL4VLM, which relies solely on outcome rewards, struggles to provide effective guidance, with both win rates and returns declining, exhibiting signs of thought collapse. In contrast, with thought guidance in GTR, although the corrector model may not be as precise as an expert policy, it is sufficient to stimulate the agent's reasoning ability. Table \ref{tab:evaluation_alfworld} lists the success rates of different methods on tasks in ALFWorld. It is worth noting that using the textual descriptions of scenes in ALFWorld as inputs significantly reduces the difficulty of recognition and decision-making, as proved by the heavy reliance observed in the output thoughts. This creates an unreasonable advantage over our setting. Nevertheless, GTR can still achieve competitive success rates through reinforcement learning.

\begin{table}[t]
\centering
    \resizebox{\linewidth}{!}{
    \begin{tabular}{cccccccccc}\toprule
     \multirow{2}{*}{\textbf{Model}} & \multirow{2}{*}{\textbf{Size}} & \multicolumn{2}{c}{\textbf{Numberline}} & \multicolumn{2}{c}{\textbf{EZPoints}} & \multicolumn{2}{c}{\textbf{Blackjack}} \\
      & & \textbf{SR(\%)}  & \textbf{ER} & \textbf{SR(\%)}  & \textbf{ER} & \textbf{SR(\%)}  & \textbf{ER}  \\ \midrule
    CNN+RL* & / & 87.1 & 0.79 & 0 & -1.02 & 38.8 & -0.17 \\
    Gemini* & API & 82.5 & 0.74 & 2.0 & -2.57 & 30.0 & -0.35 \\
    GPT4-V* & API & 65.5 & -0.59 & 10.5 & -1.30 & 25.5 & -0.44 \\
    GPT4o & API & \textbf{100.0} & \textbf{1.00} & 79.0 & 7.0 & 36.0 & -0.19 \\
    Qwen2-VL* & 72B & \textbf{100.0} & / & \textbf{100.0} & / & \textbf{42.6} & / \\
    LLaVA-sft & 7B & 59.5 & -2.61 & 39.0 & 0.67 & 25.5 & -0.46 \\ 
    RL4VLM & 7B & 90.5 & 0.89 & 48.0 & 4.19 & 40.1 & -0.16 \\
    \textbf{GTR} & 7B & \textbf{100.0} & \textbf{1.00} & 94.5 & \textbf{9.43} & 41.3 & \textbf{-0.11} \\ \bottomrule
    \end{tabular}
    }
    \caption{\textbf{Performance of different models in other tasks in $\mathtt{gym\_cards}$.} On simpler tasks where thought collapse is not evident, GTR still achieves improvements over RL4VLM and is comparable to the pretrained model 10x larger. SR - success rate, ER - episode return, * - reported in previous work.}
    \label{tab:evaluation_other}
\end{table}

\begin{figure*}[t]
\begin{floatrow}
\ffigbox[0.45\textwidth]{%
\centering
\includegraphics[width=0.9\linewidth]{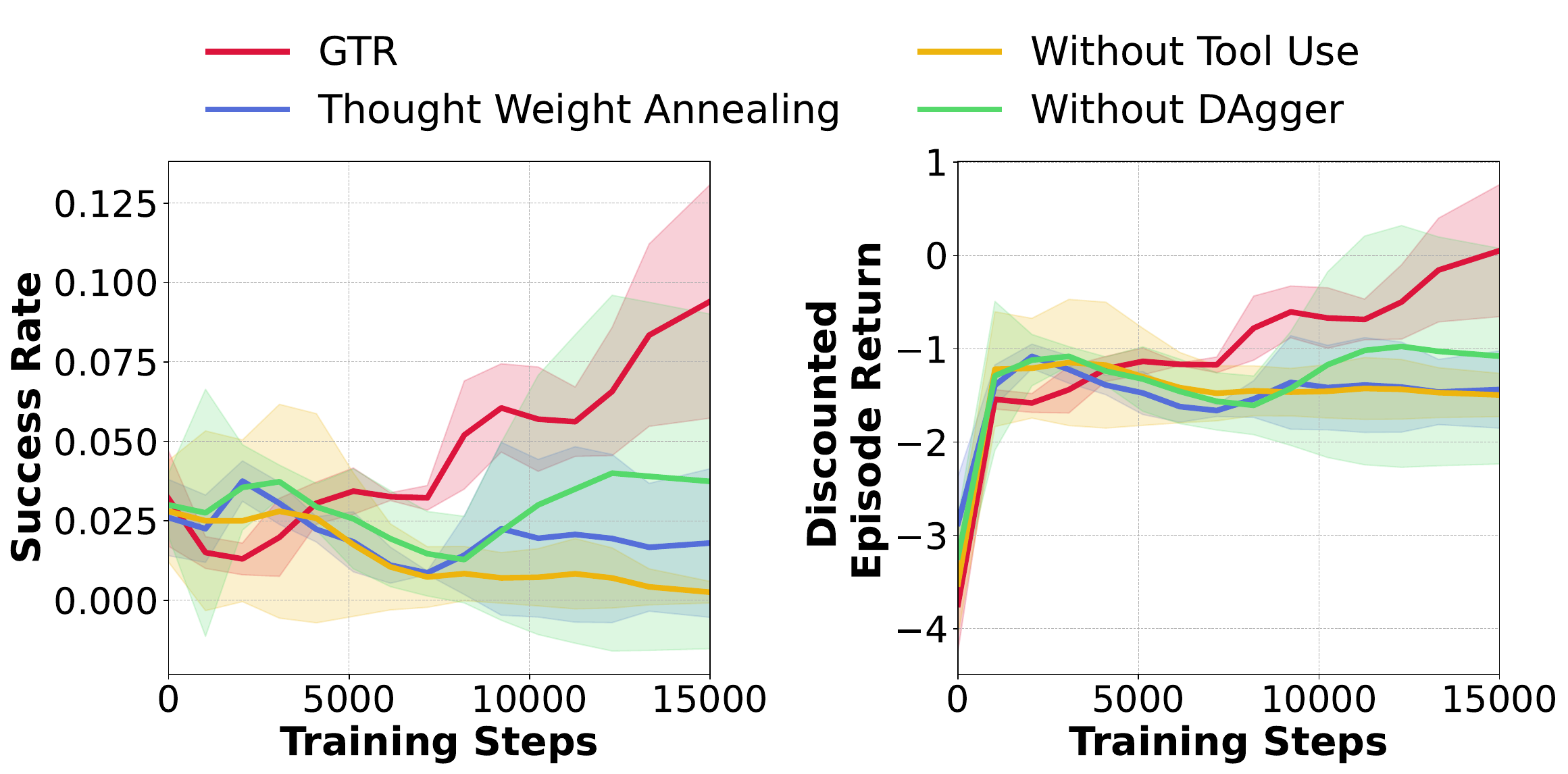}
}{%
  \caption{\textbf{Ablation study of the GTR framework.} Its superior performance highlights the importance of full-process thought guidance, task-specific knowledge, and DAgger.}
  \label{fig:ablation_gtr} 
}
\ffigbox[0.5\textwidth]{%
\includegraphics[width=0.9\linewidth]{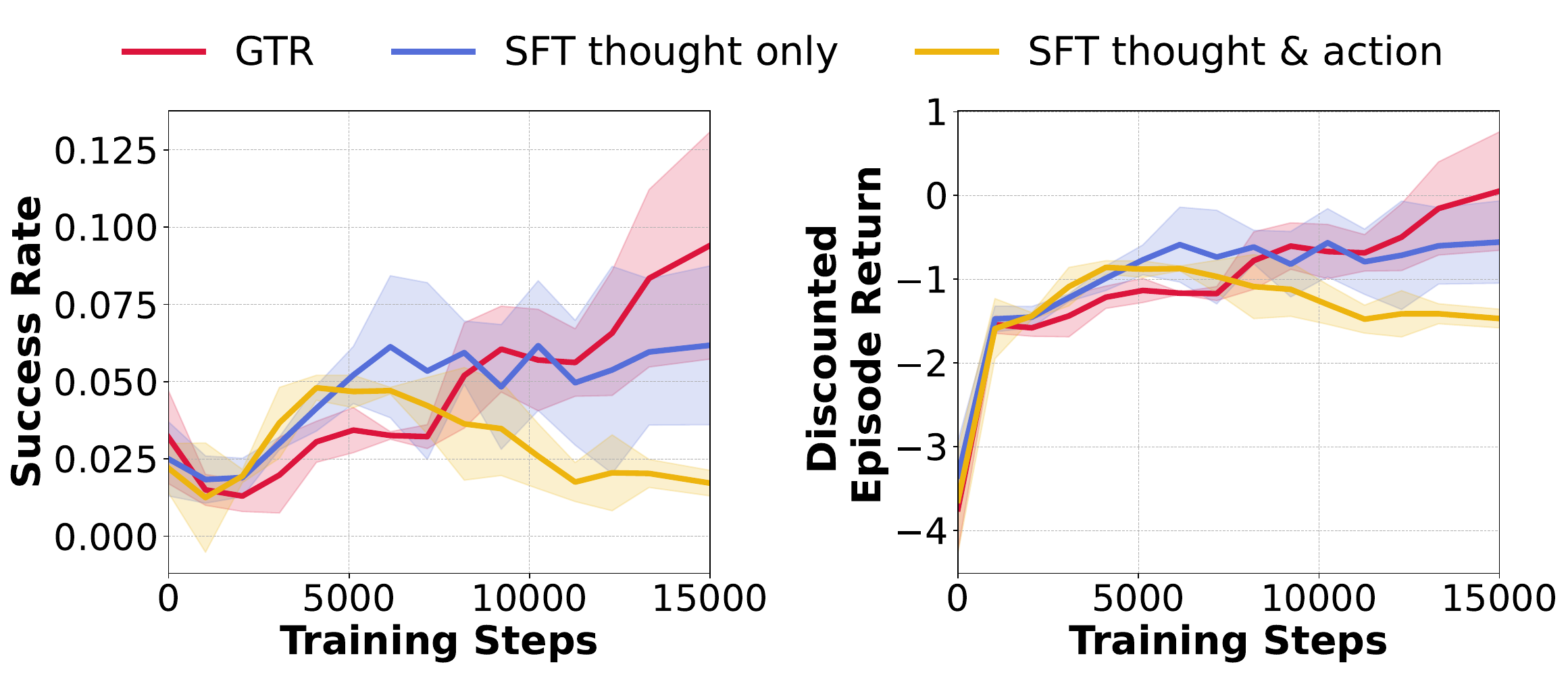}
}{%
  \caption{\textbf{Comparing thought cloning with SFT on both thoughts and actions.} This full response cloning approach does not yield satisfactory results due to its vulnerability to the corrector's hallucinations.}
  \label{fig:ablation_sft} 
}
\end{floatrow}
\vspace{-5pt}
\end{figure*}

\subsection{Ablation Study}

\paragraph{The necessity of process guidance throughout training} We aim to investigate whether adjusting the loss weights can enable a training regime where the agent imitates reasoning early on and focuses on free exploration later. To this end, we apply cosine annealing to the weight of the thought cloning loss, gradually reducing its proportion in the total loss function. As shown in Figure \ref{fig:ablation_gtr}, the training curve with annealing fails to achieve the same breakthrough as GTR. A closer examination of the model's outputs reveals that relaxing process guidance may lead to a return to thought collapse. This fact proves that process guidance is essential throughout the training process to maintain the stability and effectiveness of RL training.

\paragraph{The importance of a capable corrector model} We experiment with removing the tool function calling ability from the GPT-4o corrector, which significantly impairs its ability to analyze and solve the 24-point game task. In this setup, the reference thought provided to the agent lacks rationality, and performance did not improve, which aligns with intuition. Although the agent retained some visual capabilities, its reasoning remained illogical and disconnected from action decisions. This demonstrates that the corrector model must possess sufficient analytical and problem-solving abilities to synergize effectively with RL training. It also highlights the value of incorporating task-specific knowledge when using general-purpose foundation models as correctors.

\paragraph{GTR benefits from DAgger} In Figure \ref{fig:ablation_gtr}, we compare the training curves of GTR with and without DAgger. Results show that removing DAgger still allows the model to achieve increased performance in the early stage, but further breakthroughs become difficult as training progresses. This indicates that distribution shift and forgetting caused by the evolving thought cloning dataset indeed hinder the continual improvement of RL learning. This issue is alleviated by adopting the DAgger method from imitation learning, which constantly expands the dataset for thought cloning.


\paragraph{Thought cloning vs. full response cloning} We also try performing SFT on both thoughts and actions. Figure \ref{fig:ablation_sft} shows that SFT on the full response does not yield promising results. From detailed outputs, we find instances where mismatches between thoughts and actions are validated due to corrector hallucinations. We conclude that the constraints on actions from the environment and corrector interfere, and the stronger SFT constraints make the agent more susceptible to corrector hallucinations, undermining its ability to adjust based on environmental feedback. Cloning thoughts alone remains a more balanced and robust approach.



\section{Conclusion}
\label{sec:conclusion}

During the RL finetuning of VLM agents for challenging tasks, we identified the thought collapse issue, where the lack of thought supervision leads the agent to generate state-irrelevant reasoning, ultimately losing its ability to think coherently. Therefore, we propose the Guided Thought Reinforcement framework, which introduces an automated corrector to refine the agent's reasoning on the fly. This simple and scalable approach provides effective process guidance. Combining thought reinforcement with the well-established RL framework, GTR unleashes the agent's decision-making capabilities, enabling significantly smaller models to achieve substantial advantages in complex, long-horizon tasks. 

While our study suggests that process guidance is effective for RL training of VLM agents in multi-step tasks, we have not explored the approach promoting o1-like long CoT for action sequence reasoning, which remains an interesting direction for future research. Additionally, due to resource limitations, our research primarily focuses on 7B-scale models. Larger models may further enhance agent performance.
{
    \small
    \bibliographystyle{ieeenat_fullname}
    \bibliography{main}
}

\newpage
\appendix
\onecolumn
\section{Additional Details on Environments}
We provide a detailed introduction to the experimental environments used in this study.

\subsection{Points24}
\paragraph{State and action space.} At each state $s_t$ in the Points24 task, the agent observes an image showing four poker cards and a text-based representation of the current formula. The goal is to form a formula equal to 24 using the numbers represented by the four cards and basic operators. Card ``J'', ``Q'', ``K'' are all treated as number 10. The action space includes \{``1'', ``2'', $\ldots$, ``10'', ``+'', ``-'', ``*'', ``/'', ``('', ``)'', ``=''\}, and each card can only be used once. Selecting a number not present in the image or one that has already been used is considered an illegal action. If the action is legal, the corresponding number or operator is appended to the current formula, forming the next state $s_{t+1}$; if the action is illegal, the state remains unchanged $s_{t+1}=s_t$. The environment does not guarantee that the four cards in the image have a feasible solution equal to 24.

\paragraph{Reward function.} At each step, the agent receives a reward $r=-1$ for outputting an illegal action and a reward $r=0$ for a legal action. The episode terminates when the agent outputs ``='' as an action or the step counts exceeds $T=20$. At termination, if the formula evaluates to 24, the agent receives an outcome reward $r=10$; otherwise, it receives $r=-1$.

\begin{figure}[ht]
  \centering
    \includegraphics[width=0.75\linewidth]{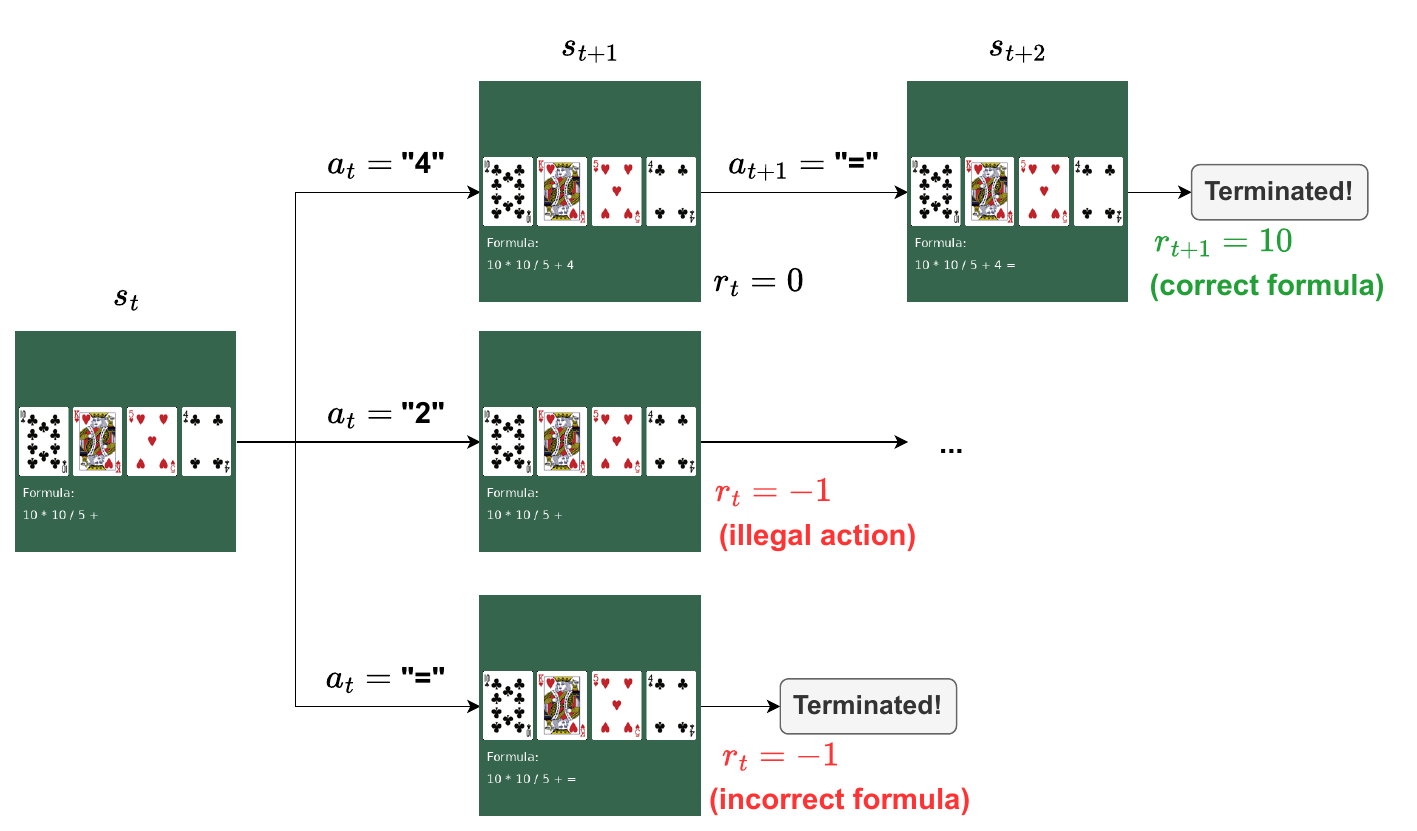}
  \caption{\textbf{The Points24 task.}}   
\end{figure}

\subsection{ALFWorld}
\paragraph{State and action space.} In the ALFWorld environment in our experiments, the agent receives an RGB observation image and a history of past actions at each state $s_{t}$. The action space includes all possible interactions in the current scenario, typically categorized as: (1) $\mathtt{go\ to\ \{recep\}}$, (2) $\mathtt{take\ \{obj\} \ from\ \{recep\}}$, (3) $\mathtt{put \ \{obj\} \ in/on\ \{recep\}}$, (4) $\mathtt{open\ \{recep\}}$, (5) $\mathtt{close\ \{recep\}}$, (6) $\mathtt{toggle\ \{obj\}\ \{recep\}}$, (7) $\mathtt{clean\ \{obj\} \ with\ \{recep\}}$, (2) $\mathtt{heat\ \{obj\} \ with\ \{recep\}}$, (2) $\mathtt{cool\ \{obj\} \ with\ \{recep\}}$, where $\mathtt{\{obj\}}$ and $\mathtt{\{recep\}}$ denote objects and receptacles. After an admissible action is taken, ALFWorld renders the updated scene from the agent's view as the next state $s_{t+1}$. $s_{t+1}=s_t$ if the action is illegal.

Notably, the original ALFWorld environment provides a text description of the scene in each state without the action history. However, to prevent the agent from relying on the textual description rather than visual observation and to better simulate real-world scenarios, we modified the state by removing the text description and adding the action sequence taken. This adjustment increases the difficulty, emphasizing the agent's visual recognition and long-horizon decision-making capabilities.

\paragraph{Reward function.} The reward system of ALFWorld consists of two components. Each state $s$ has a set of admissible actions $\mathcal{A}_\text{adm}(s)$, and illegal actions are penalized. Additionally, each task in ALFWorld has both the final goal $g_\text{task}$ and sub-goals $g_\text{sub}$, and achieving these goals also provides rewards. Formally, the reward function can be written as:
\begin{equation}
    r(s_t, a_t, s_{t+1}|g_\text{task}) = 50 \times \mathbf{1}(s_{t+1} = g_\text{task}) + \mathbf{1}(s_{t+1} = g_\text{sub}) - \mathbf{1}(a_t \notin \mathcal{A}_\text{adm}(s)).
\end{equation}

\begin{figure}[ht]
  \centering
    \includegraphics[width=\linewidth]{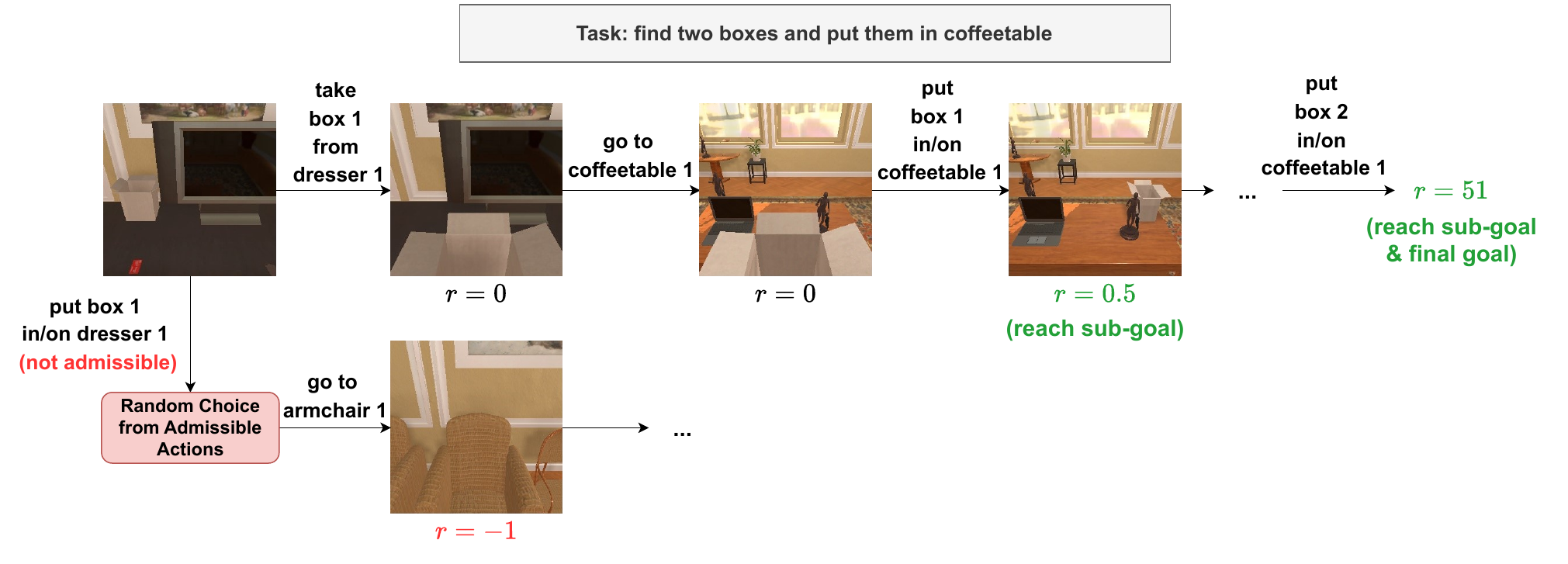}
  \caption{\textbf{The ALFWorld task.}}   
\end{figure}

\subsection{Other Games in the \textit{gym\_cards} Environment}
We briefly introduce the other tasks in the $\mathtt{gym\_cards}$ environment, which have significantly smaller state and action spaces, shorter episode lengths, and lower complexity than the two tasks we selected. As a result, these tasks do not exhibit the thought collapse phenomenon. Nevertheless, GTR still achieves performance improvements in these more straightforward tasks.

\paragraph{Numberline.} The agent receives an image displaying the text ``Target: $x$'' and ``Current: $y$'', where $x,y$ are integers in $[0,5]$. The action space is \{+, -\}, which increments or decrements the current number by 1, respectively. The goal is to make the current number equal to the target number. The agent gets a reward of 1 upon achieving the goal and a penalty of -1 if an action moves the current number away from the target. The game can always be solved within 5 steps.

\paragraph{EZPoints.} This task is a simplified variant of Points24, with the image containing only two cards, the available operators limited to \{+, -, =\}, and the target value is 12. In addition, the EZPoints environment guarantees that the two cards in the image always have a valid solution. The correct formula always takes 4 steps.

\paragraph{Blackjack.} The task is to win the blackjack game. The image at each state includes two cards of the dealer (one of them facing down) and all cards from the player. The action space is \{``stand'', ``hit''\}. The agent gets one more card when choosing ``hit'', and the game terminates when choosing ``stand''. Theoretically, the player has an expected winning rate slightly below 50\%.

\begin{figure}[ht]
  \centering
    \includegraphics[width=0.7\linewidth]{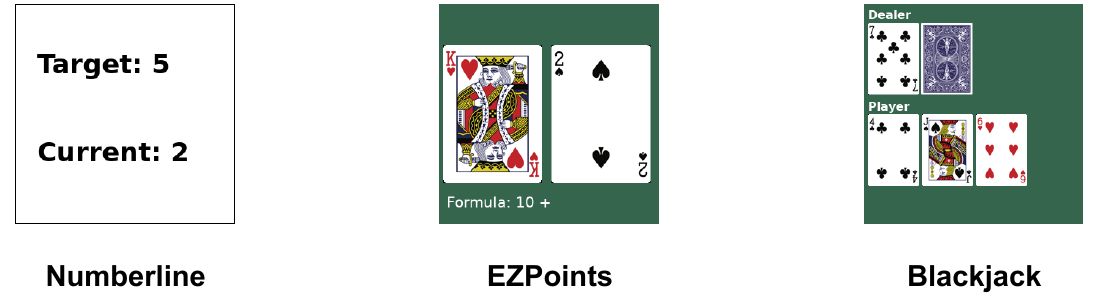}
  \caption{\textbf{Other tasks in the $\texttt{gym\_cards}$ environment.}}   
\end{figure}

\section{Additional Details on Training}
\subsection{Training Setting}
Drawing inspiration from the RLHF training framework \cite{ouyang2022training} and prior related work \cite{zhai2025fine}, we perform one epoch of supervised fine-tuning on the base LLaVA-v1.6-mistral-7B model \cite{liu2023llava, liu2023improvedllava, liu2024llavanext} before RL training, which is referred to as \textit{LLaVA-sft} in the results. The datasets are sourced from the RL4VLM paper \cite{zhai2025fine}, with labels for the $\mathtt{gym\_cards}$ environment provided by a task solver and labels for the ALFWorld environment generated by GPT-4V. 

\subsection{Hyperparameters}
In Table \ref{tab:hyperparameters}, we provide the hyperparameter settings used for GTR training, which are primarily derived from values proposed in previous work \cite{zhai2025fine}. We employ LoRA \cite{hu2022lora} to fine-tune the entire VLM model, including the CLIP vision encoder \cite{radford2021learning}, LLM backbone, and MLP projector, enabling it to run on an A100 GPU with 40GB memory.

\begin{table}[htb]
\caption{\textbf{Hyperparameters of GTR}}
\label{tab:hyperparameters}
\begin{center}
\begin{small}
\begin{tabular}{lcc}
\toprule
\textbf{Hyperparameter} & & \textbf{Value}  \\
\midrule
\multicolumn{3}{c}{\textbf{General Setup - Training}} \\
Learning rate     && $\mathtt{CosineAnnealingLR}$ \\
Initial learning rate     && $1e-5$ \\
Final learning rate    && $1e-9$ \\
Maximum learning rate step  && $25$ \\
Discount factor $\gamma$  && $0.9$ \\
GAE $\lambda$ && $0.95$ \\
PPO entropy coefficient && $0.01$ \\
PPO value loss coefficient && $0.5$ \\
PPO clip parameter $c$ && $0.1$ \\
PPO epoch && $4$ \\
Gradient accumulation steps && $128$ \\
LoRA $r$ && $128$ \\
LoRA $\alpha$ && $256$ \\
LoRA $\mathtt{dropout}$ && $0.05$ \\
\midrule
\multicolumn{3}{c}{\textbf{General Setup - Models}} \\
Generation max text length && $256$ \\
Generation temperature && $0.2$ \\
Generation repetition penalty && $1.2$ \\
Corrector max text length  && $600$ \\
Corrector temperature  && $0.4$ \\
\midrule
\multicolumn{3}{c}{\textbf{For Points24 task}} \\
Environmental steps && $15000$ \\
Thought probability coefficient && $0.5$ \\
\midrule
\multicolumn{3}{c}{\textbf{For ALFWorld task}} \\
Environmental steps && $5000$ \\
Thought probability coefficient && $0.2$ \\
\bottomrule
\end{tabular}
\end{small}
\end{center}
\end{table}

\subsection{Computational Overhead}
Running a large corrector model at each RL step introduces additional computational overhead during training. To provide a more comprehensive comparison of corrector models of different types and sizes, we evaluate the performance, cost, and training time of both open-source and closed-source models, as well as large and small models. The results in Table \ref{tab:ablation_overhead} show that although open-source models can reduce overhead, the performance of Qwen2.5-VL-72B is undermined by its sub-optimal tool-use capabilities, and the 7B version fails to follow proper correction formats.

\begin{table}[htb]
    \centering
    \vspace{-1em}
    \begin{tabular}{cccc}\toprule
    \textbf{Corrector Model} & \textbf{Performance} & \textbf{Token Usage (Cost)}  & \textbf{Time}  \\ \midrule
    GPT-4o & 17.5\% & 33.5M (\textasciitilde\$463.5) & 86h \\
    Qwen2.5-VL-72B & 6.5\% & 33.8M (\textasciitilde\$91.6) & 110h \\
    Qwen2.5-VL-7B & N/A & 31.2M (\textasciitilde\$19.9) & 56h \\ \bottomrule
    \end{tabular}
    \caption{\textbf{Ablation study on computational overhead across different corrector models.}}
  \label{tab:ablation_overhead}
\end{table}

\section{Additional results of ALFWorld with textual observation}
We observe that RL4VLM's performance on ALFWorld in its original paper is attributed to precise textual observations, which notably compensates for the agent’s limited visual recognition and reasoning capabilities. This is also the primary reason why we remove the text description in our experiments. Nevertheless, we include the performance of RL4VLM and GTR with the presence of textual observations. The results in Table \ref{tab:alfworld_with_text} demonstrate that GTR does not need textual observations but achieves competitive performance as the corrector effectively bridges the gap between vision and text.

\begin{table}[htb]
    \centering
    \vspace{1em}
    \begin{tabular}{ccc}\toprule
    \textbf{} & \textbf{Success Rate}  \\ \midrule
    \cellcolor{cyan!10}RL4VLM w/ text & \cellcolor{cyan!10}21.7\% \\
    \cellcolor{cyan!10}RL4VLM w/o text & \cellcolor{cyan!10}5.4\% \\
    \cellcolor{pink!20}GTR w/ text & \cellcolor{pink!20}21.0\% \\
    \cellcolor{pink!20}GTR w/o text & \cellcolor{pink!20}17.8\% \\ \bottomrule
    \end{tabular}
    \caption{\textbf{Performance comparison of ALFWorld with and without textual observations.}}
    \label{tab:alfworld_with_text}
\end{table}

\section{Continual Training of GTR}
\begin{figure}[ht]
  \centering
    \includegraphics[width=0.7\linewidth]{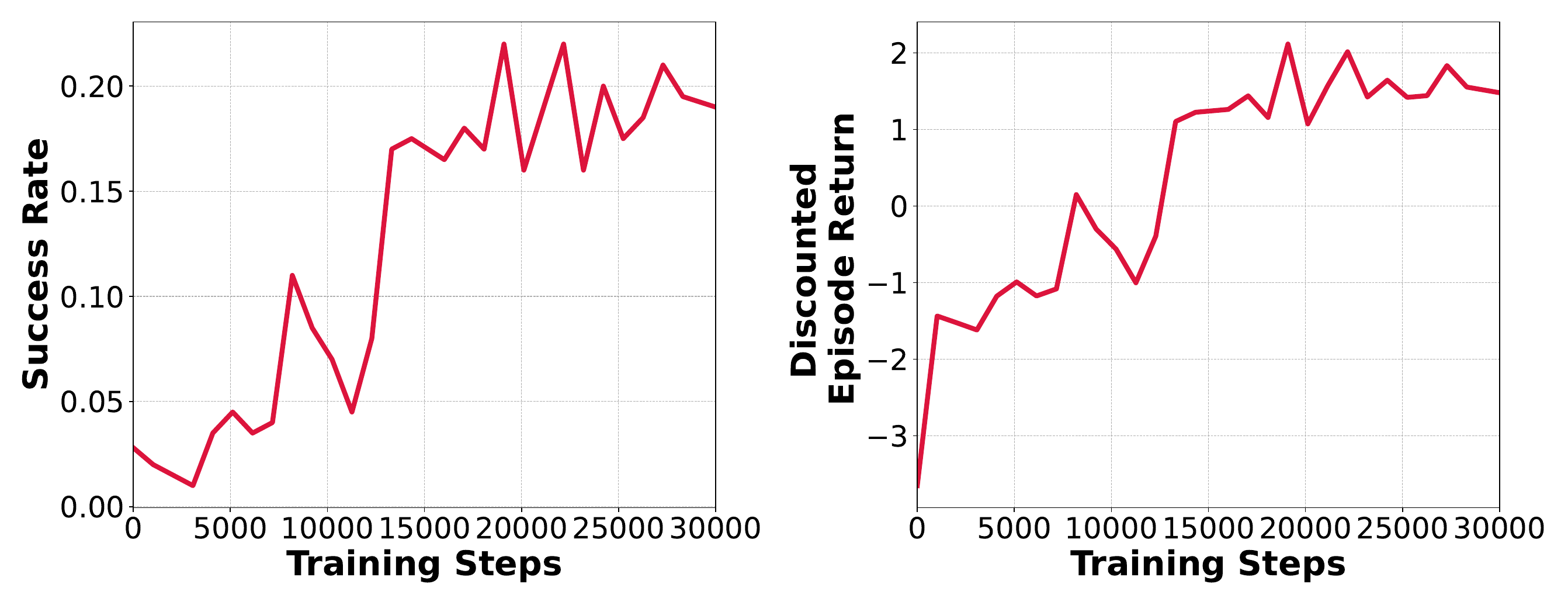}
  \caption{\textbf{Training the VLM agent with GTR for 30,000 steps.}}   
  \label{fig:gtrextend}
\end{figure}

To evaluate the performance of the GTR algorithm over extended training durations, we present results of GTR trained for 30k steps on the Points24 task. As shown in Figure \ref{fig:gtrextend}, GTR is able to maintain excellent performance.

\newpage
\section{Prompts of the Corrector Model}
\begin{table*}[htb]
\scriptsize
\begin{minipage}{\linewidth}
Prompt adopted by the VLM corrector model for the Points24 task
\centering
\ttfamily
\begin{tabular}{p\linewidth}
\midrule
\textbf{System Prompt:} You are an expert 24-point card game player. You are observing four cards in the image and the current formula. The goal is to output a formula that evaluates to 24, and each number can only be used once. The number or operator include ['1', '2', '3', '4', '5', '6', '7', '8', '9', '10', '+', '-', '*', '/', '(', ')', '='], and the chosen number or operator will be appended to the current formula to reach the correct target formula that evaluates to 24. Note that 'J', 'Q', and 'K' count as '10'. \\
\textbf{Query:} You will be given the current formula, the thought of a player playing this game, and a target formula. The player's thought may be wrong, please evaluate its correctness by the following aspects: \\
(1) What are the four cards in the image? If the target formula is 'NOT DETERMINED', use the 'find\_all\_correct\_formulas' tool function to find all possible correct formulas by the four cards in the image. Remember the correct formulas, and do not output the result.\\
(2) What are the recognized card ranks in the thought? According to the rules, does the ranks in the thought match your observation in question (1), regardless of the order?\\
(3) What is the proposed formula the player is trying to reach in the thought? Does the proposed formula match the target formula or, if the target formula is 'NOT DETERMINED', one of the possible correct formulas in question (1)?\\
(4) Does the player choose the correct action to reach the proposed formula or choose '=' if the current formula is complete? \\
Please briefly answer the above questions, then give your final evaluation. If the thought is incorrect, use all available information for thought correction: determine the next single number or character to append to the current formula and finally provide the correct thought. \\
Your response should be a valid json file in the following format: \{\\
"answer1": \{Text, answer to the first question\}, \\
"answer2": \{Text, answer to the second question\}, \\
"answer3": \{Text, answer to the third question\}, \\
"answer4": \{Text, answer to the third question\}, \\
"evaluation": \{YES or NO\}, \\
"possible\_solution": \{YES or NO, indicating whether there is a possible solution. None if the thought is correct\}, \\
"target\_formula": \{The given target formula if it is not None. The proposed formula in the thought if the thought is correct. Otherwise, choose an appropriate target formula from all possible correct formulas obtained from the tool function for the player to reach. \},\\
"correction": \{Json object, the correct thought. None if the thought is correct\} \\
\} \\
\text{[Current Formula]} ... \\
\text{[Thought]} ... \\
\text{[Target Formula]} ... \\
\midrule
\end{tabular}
\end{minipage}%
\label{tab:prompt_points24}
\end{table*}

\begin{table*}[htb]
\scriptsize
\begin{minipage}{\linewidth}
Prompt adopted by the VLM corrector model for the ALFWorld task
\centering
\ttfamily
\begin{tabular}{p\linewidth}
\midrule
\textbf{System Prompt:} You are an expert in the ALFRED Embodied Environment. The environment requires the player to navigate, take certain objects, interact with objects if necessary, and finally put objects in the designated place to complete the task. \\
\textbf{Query:} You will be given the visual observation and thought of a player in this environment. The task is to ... You are also given the previous actions the player has taken: ... All admissible actions of the current situation are: ... \\
Please evaluate if the reasoning is correct by the following aspects:\\
(1) What objects are in your sight and whether you are holding a certain object? Does the thought correctly identify the image?\\
(2) Based on the task description and the action history, what should be the player's next sub-goal (notice that the tasks require the player to first pick up certain objects, interact with receptacles if the task is cooling, heating, cleaning or looking in light, and finally placing the object)? Does the thought align with the sub-goal?\\
(3) Based on the task description and the action history, does the player choose one of the admissible actions to reach the sub-goal? Does the action take effect? If the target object is not in sight, go to an unexplored location; if there is a required object, take it; if the task requires cooling, heating, cleaning, or looking in light, navigate and interact with the receptacles.\\
Please briefly answer the above questions, then give your final evaluation. If the thought is incorrect, include all available information for thought correction: choose one correct step from the admissible actions for the player to finish the task, and finally provide the correct thought. \\
Your response should be a valid json file in the following format: \{\\
"answer1": \{Text, answer to the first question\}, \\
"answer2": \{Text, answer to the second question\}, \\
"answer3": \{Text, answer to the third question\}, \\
"evaluation": \{YES or NO\}, \\
"correction": \{Json object, the correct thought. None if the thought is correct\} \\
\}\\
\text{[The player's thought]} ...\\
\midrule
\end{tabular}
\end{minipage}%
\label{tab:prompt_alfworld}
\end{table*}

\newpage
\section{Example Trajectories of Thought Collapse and GTR}
In this section, we present real episode examples from the Points24 and ALFWorld tasks that exhibit thought collapse during RL training. These examples illustrate how the model's outputs deteriorate, becoming state-irrelevant and irrational, thereby significantly impairing the agent's decision-making capabilities. This catastrophic phenomenon is the core issue addressed in our work. We also show the GTR-generated trajectories that effectively solve the problem.

\subsection{Points24}
\begin{figure}[ht]
  \centering
    \includegraphics[width=0.9\linewidth]{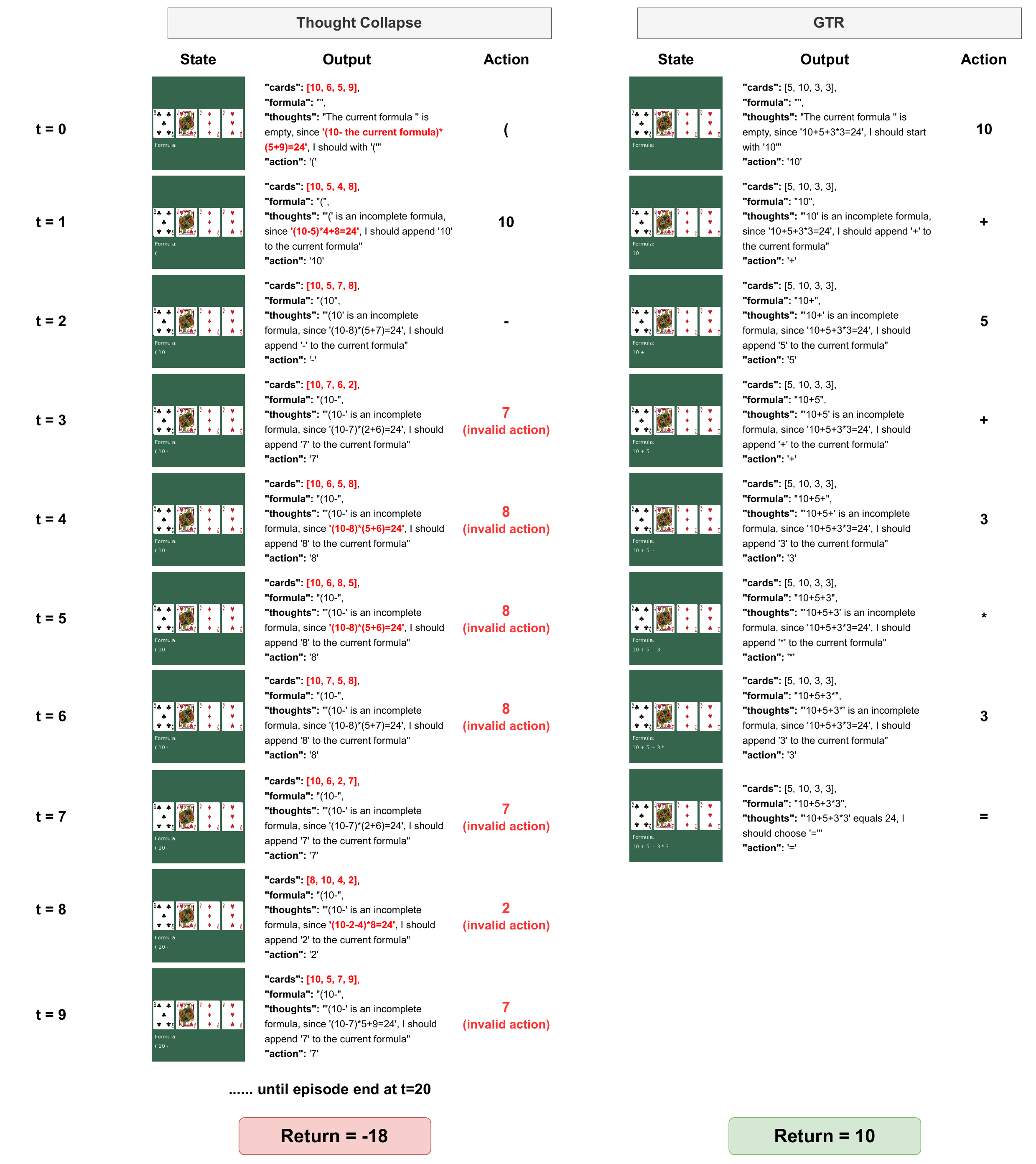}
  \caption{\textbf{Examples of agent trajectory produced by RL4VLM (left) and GTR (right) in Points24.}}   
\end{figure}

\subsection{ALFWorld}
\begin{figure}[ht]
  \centering
    \includegraphics[width=0.95\linewidth]{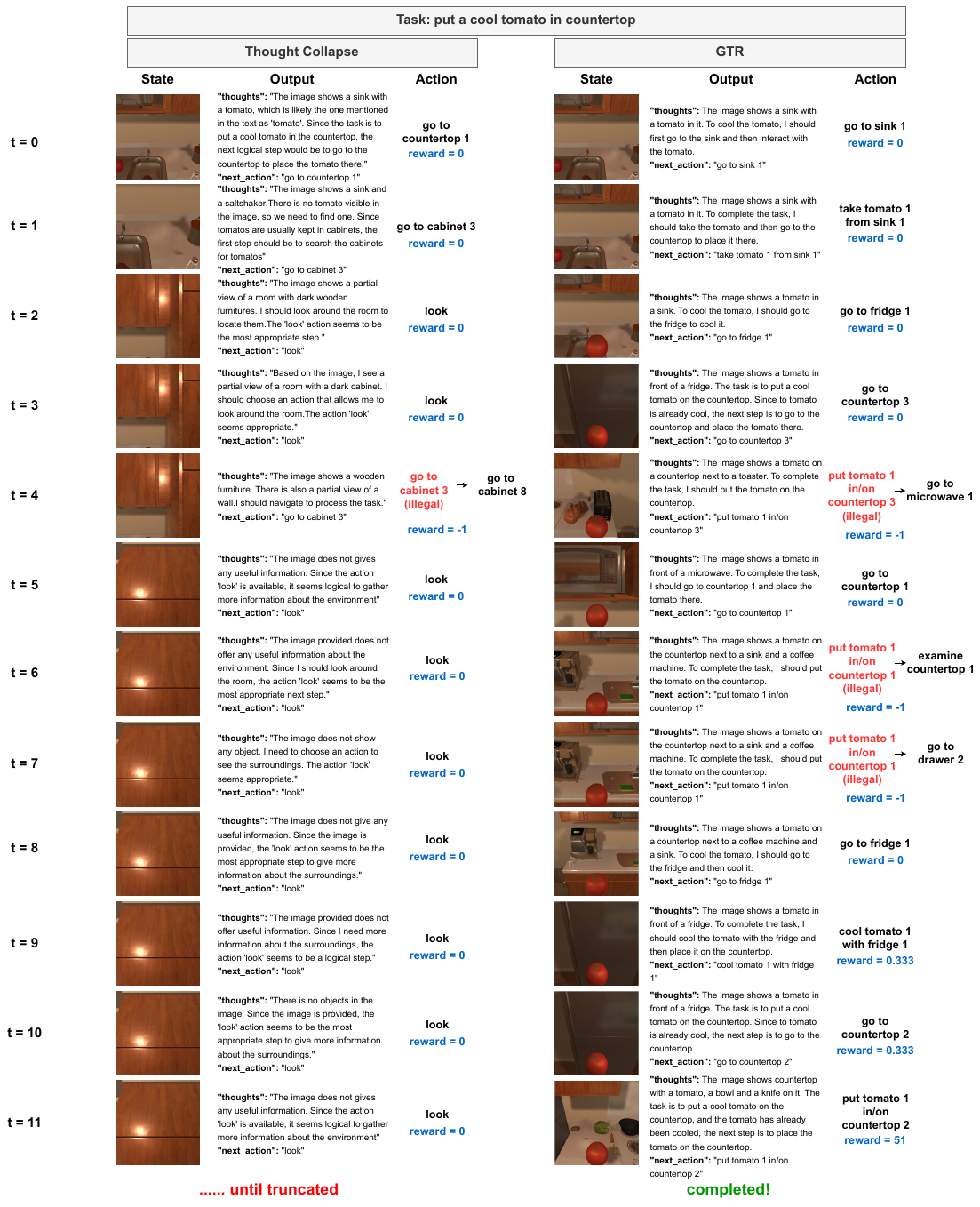}
  \caption{\textbf{Examples of agent trajectory produced by RL4VLM (left) and GTR (right) in ALFWorld.}}   
\end{figure}

\end{document}